\documentclass[conference, final]{IEEEtran}
\IEEEoverridecommandlockouts

\usepackage[table]{xcolor}

\usepackage[]{todonotes}
\presetkeys{todonotes}{inline}{}
\usepackage{glossaries}
\usepackage{savesym}
\savesymbol{checkmark}
\usepackage{dingbat}
\usepackage{bbm}
\usepackage{array, makecell}
\usepackage{tikz}
\usepackage{pgfplots}

\makeatletter
\let\MYcaption\@makecaption
\makeatother

\usepackage[font=footnotesize]{subcaption}

\makeatletter
\let\@makecaption\MYcaption
\makeatother

\usepackage[utf8]{inputenc}
\usepackage[T1]{fontenc}

\usepackage{algorithm}
\usepackage{algpseudocode}
\usepackage{icomma}

\usepackage[backend=biber,style=ieee,natbib=true]{biblatex}
\usepackage{booktabs}
\usepackage[draft,bookmarks=false]{hyperref}

\usepackage{amsfonts}
\usepackage{textcomp}

\usepackage[capitalise]{cleveref}  

\PassOptionsToPackage{table}{xcolor}
\newacronym{dfs}
{DFS}
{Depth-First Search}

\newacronym{bfs}
{BFS}
{Breadth-First Search}

\newacronym{rne}
{RNE}
{Road Network Embedding}

\newacronym{mic}
{MIC}
{Maximal Information Coefficient}

\newacronym{tsne}
{t-SNE}
{t-distributed Stochastic Neighbor Embedding}

\newacronym{etl}
{ETL}
{Extract-Transform-Load}

\newacronym{rss}
{RSS}
{Residual Sum of Squares}

\newacronym{gps}
{GPS}
{Global Positioning System}

\newacronym{pu}
{P\&U}
{Pre-Train \& Update}

\newacronym{osm}
{OSM}
{OpenStreetMap}

\newacronym{nlp}
{NLP}
{Natural Language Processing}

\newacronym{cnn}
{CNN}
{Convolutional Neural Network}

\newacronym{cane}
{CANE}
{Context-Aware Network Embedding}

\newacronym{vldb}
{VLDB}
{International Conference on Very Large Data Bases}

\newacronym{jvldb}
{VLDBJ}
{The International Journal on Very Large Data Bases}

\newacronym{cikm}
{CIKM}
{The International Conference on Information and Knowledge Management}

\newacronym{geoinf}
{GeoInf}
{GeoInformatica}

\newacronym{sigmod}
{SIGMOD}
{ACM Special Interest Group on Management of Data}

\newacronym{icde}
{ICDE}
{International Conference on Data Engineering}

\newacronym{edbt}
{EDBT}
{International Conference on Extending Database Technology}

\newacronym{mdm}
{MDM}
{IEEE International Conference on Mobile Data Management}

\newacronym{sstd}
{SSTD}
{Symposium on Spatial and Temporal Databases}

\newacronym{iclr}
{ICLR}
{International Conference on Learning Representations}

\newacronym{tkde}
{TKDE}
{ACM Transactions on Knowledge and Data Engineering}

\newacronym{tods}
{TODS}
{ACM Transactions on Database Systems}

\newacronym{tsas}
{TSAS}
{ACM Transactions on Spatial Algorithms and Systems}

\newacronym{sigkdd}
{SIGKDD}
{ACM Special Interest Group on Knowledge Discovery and Data Mining}

\newacronym{tnnis}
{TNNIS}
{IEEE Transactions on Neural Networks and Learning Systems}

\newacronym{aaai}
{AAAI}
{The Assocation for the Advancement of Artifical Intelligence}

\newacronym{nips}
{NIPS}
{Neural Information Processing Systems}

\newcommand{\COMMENT}[1]{%
}

\begin{document}

\title{On Network Embedding for
\\ Machine Learning on Road Networks:
\\ A Case Study on the Danish Road Network}

\author{
  \IEEEauthorblockN{
    Tobias Skovgaard Jepsen,
    Christian S. Jensen,
    Thomas Dyhre Nielsen,
    Kristian Torp
  }
  \IEEEauthorblockA{
    Department of Computer Science \\
    Aalborg University, Aalborg, Denmark\\
    \{tsj, csj, tdn, torp\}@cs.aau.dk
  }
  \COMMENT{
  \IEEEauthorblockA{
    \IEEEauthorrefmark{2}Center for Data Intensive Systems\\
    Aalborg University, Aalborg, Denmark\\
    csj@cs.aau.dk
  }
  \IEEEauthorblockA{
    \IEEEauthorrefmark{3}Distributed, Embedded and Intelligent Systems\\
    Aalborg University, Aalborg, Denmark\\
    tdn@cs.aau.dk
  }
  \IEEEauthorblockA{
    \IEEEauthorrefmark{4}Center for Data Intensive Systems\\
    Aalborg University, Aalborg, Denmark\\
    torp@cs.aau.dk
  }
} 
}

\maketitle
\begin{abstract}
  Road networks are a type of spatial network, where edges may be associated with qualitative information such as road type and speed limit. Unfortunately, such information is often incomplete; for instance, OpenStreetMap only has speed limits for 13\% of all Danish road segments.
This is problematic for analysis tasks that rely on such information for machine learning.
To enable machine learning in such circumstances, one may consider the application of network embedding methods to extract structural information from the network.
  However, these methods have so far mostly been used in the context of social networks, which differ significantly from road networks in terms of, e.g., node degree and level of homophily (which are key to the performance of many network embedding methods).

We analyze the use of network embedding methods, specifically node2vec, for learning road segment embeddings in road networks. Due to the often limited availability of information on other relevant road characteristics, the analysis focuses on leveraging the spatial network structure. Our results suggest that network embedding methods can indeed be used for deriving relevant network features (that may, e.g, be used for predicting speed limits), but that the qualities of the embeddings differ from embeddings for social networks.
\end{abstract}

\def\IEEEkeywordsname{Keywords}

{
  \footnotesize
  \emph{\textcopyright{} 2018 IEEE.  Personal use of this material is permitted.  Permission from IEEE must be obtained for all other uses, in any current or future media, including reprinting/republishing this material for advertising or promotional purposes, creating new collective works, for resale or redistribution to servers or lists, or reuse of any copyrighted component of this work in other works.}
}

\section{Introduction}
Road networks represent an important class of spatial networks and are an essential component of modern societal infrastructure. Road networks are associated with many important analysis tasks such as traffic flow and travel pattern analyses.
In particular, many important road network tasks are supported by machine learning algorithms, including travel-time estimation~\citep{trajectory-regression,realroad2vec}, traffic forecasting~\citep{road2vec}, and $k$ nearest points-of-interest queries~\citep{Shahabi2003,Liu2011}, that require set of informative features to describe, e.g., the different road segments.

Solving road network analysis tasks is difficult since there is often little information available beyond the network structure itself.
For instance, the Danish road network from \gls{osm}~\citep{osm} contains only the network structure and up to two attributes characterizing each road segments: road category and speed limit. In addition, only $13\%$ of the road segments have a speed limit label, even when augmented with data from Danish municipalities.
This information sparsity makes it difficult to derive the features necessary for solving many road network analysis tasks.
The road network structure is a potentially rich source of information, but it is not straight-forward to capture and utilize this often highly complex structure.
For road network analyses, this typically involves explicit modeling of spatial correlations between adjacent road segments based on domain knowledge~\citep{trajectory-regression,yang2014using,traffic-forecasting}.


A road network is commonly modeled as a directed graph $G = (V, E)$, where each node $v \in V$ represents an intersection or the end of a road and each edge $(u, v) \in E$ represents a directed road segment that allows travel from $u$ to $v$.
Such graph representations makes \emph{network embedding methods}---a class of feature learning methods for graphs---directly applicable for extracting structural information from road networks.

\begin{figure}[t]
  \centering
  \begin{subfigure}{0.49\columnwidth}
    \includegraphics[width=\textwidth]{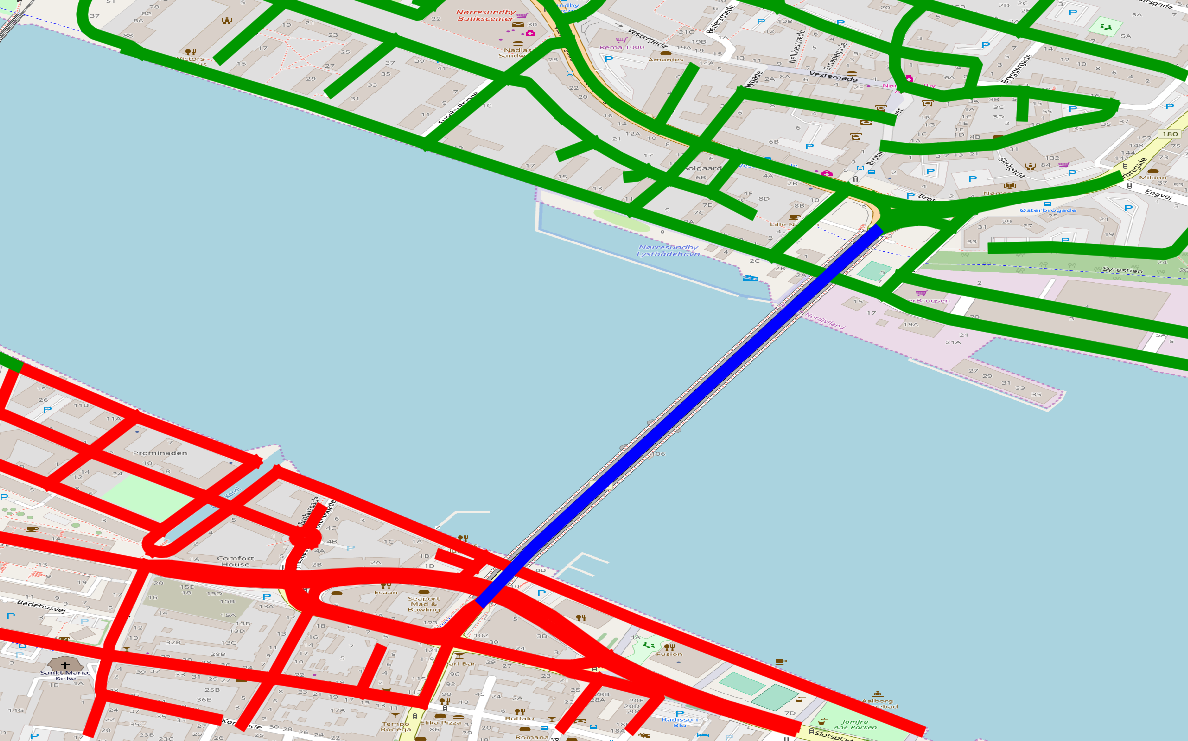}
     \caption{\label{fig:compelling-example-road-segments}}
  \end{subfigure}
  \begin{subfigure}{0.49\columnwidth}
    \includegraphics[width=\textwidth, trim={1.1cm, 0.75cm, 1.5cm, 1.21cm}, clip]{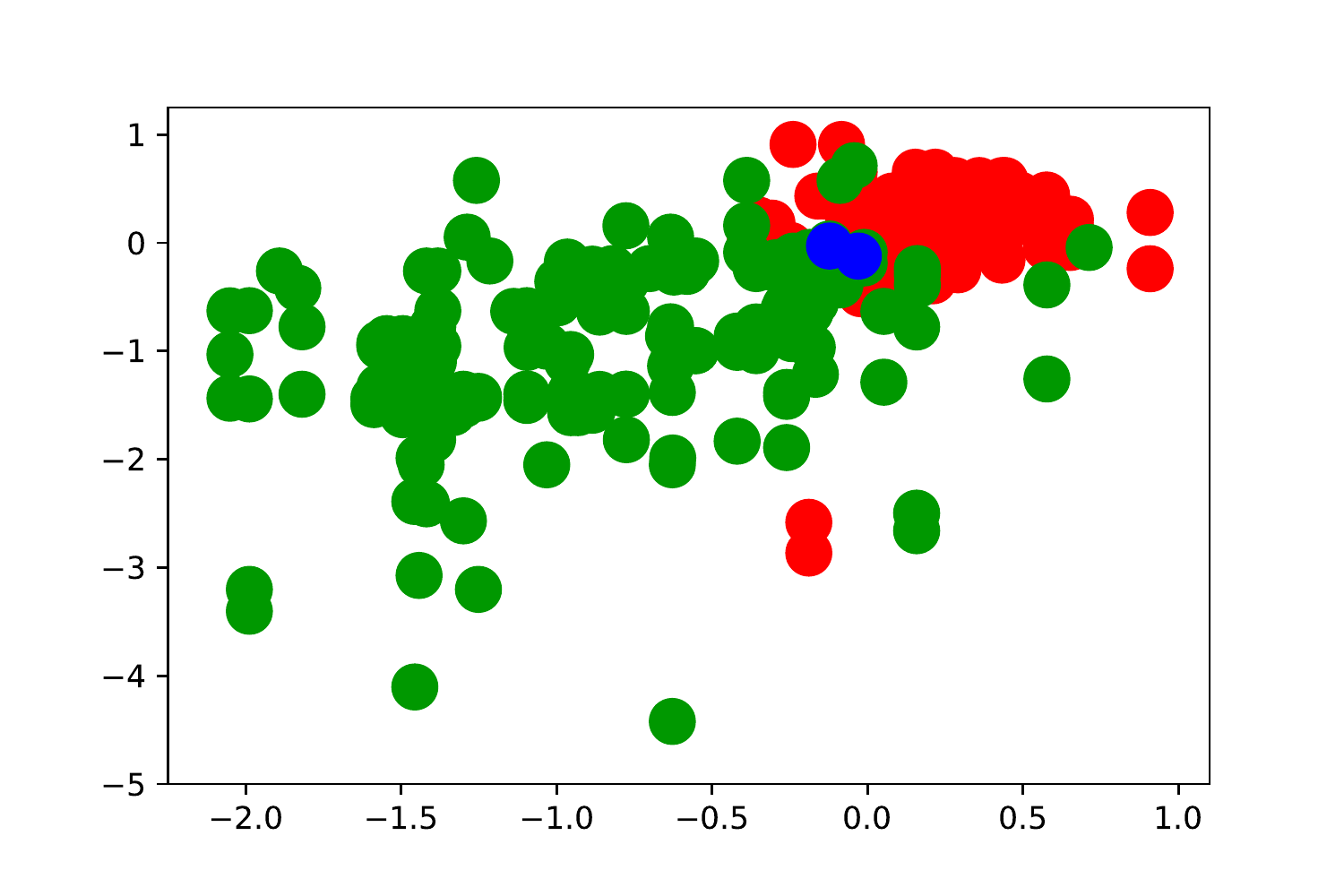}
    \caption{\label{fig:compelling-example-embedding}}
  \end{subfigure}
  \caption{Illustration of (a) road segments in the Danish road network and (b) their feature vector representation generated by \emph{DeepWalk}~\citep{deepwalk}. Colors indicate the region a road segment belongs to.\label{fig:compelling-example}}
\end{figure}

In network embedding, the goal is to learn a mapping (an \emph{embedding}) that embeds nodes in networks into a $d$-dimensional vector space s.t.\ the node neighborhoods are preserved in the embedding space~\citep{hamilton2017representation}.
In other words, nodes are mapped to feature vectors that encode the structural information of the graph s.t.\ nearby nodes in the network are mapped to vectors that are near each other in the embedding space.
For instance, \cref{fig:compelling-example-embedding} shows that road segments north and south of the bridge in \cref{fig:compelling-example-road-segments} tend to cluster with other road segments from the same region.
The road segments representing the bridge are somewhere in-between.
Network embedding methods can extract the structural information in networks to supplement or replace attribute information if such information is low-quality, sparse, or unavailable.

The research in network embedding has thus far focused primarily on social, biological, and information networks~\citep{deepwalk,node2vec,struc2vec,graphsage,lane,cane,sne,multiview}.
Such networks differ significantly from road networks in terms of, e.g., structure, semantics, size, node degree, network diameter, and the amount of attribute information available.
In addition, road networks may be disconnected due to inaccuracies in their spatial representation or the presence of islands, whereas, e.g., social networks are strongly connected.
The effect of this disconnectedness on the embeddings is not obvious.

The differences between the types of networks studied in the network embedding literature, e.g., social networks, and road networks puts into question the suitability of using network embedding methods for road networks.
We therefore formulate the following research question:
\begin{quote}
  \emph{Are existing network embedding methods suitable for performing analysis tasks on road networks?}
\end{quote}

To address this question, we conduct a case study: we evaluate an existing network embedding method empirically on road category classification and speed limit classification in the Danish road network. 
Only the road network structure is available beyond the road category and speed limit attributes which we wish to predict.
\emph{node2vec}~\citep{node2vec} is therefore our network embedding method of choice since it relies solely on network structure while optimizing for the core property of neighborhood preservation in the embedding space; a property shared by most network embedding methods.
Thus, node2vec is applicable when little or no attribute information is available in the network as is the case in our data set.
We use the node2vec algorithm to learn embeddings of road segments in the Danish road network and subsequently feed these to a classifier to predict either road categories or speed limits.

Our key contribution is an empirical evaluation of network embedding methods for solving road network analysis tasks using our case study.
In the network embedding literature, linear classifiers are commonly used to evaluate network embedding methods and can achieve high performance scores. This suggests that network embedding methods tend to create embeddings that are linearly separable relative to the classification problem at hand, a highly-desirable property that makes it easier to apply machine learning algorithms.
We see no reason to assume that these observations extend to the road segment classification tasks we consider. We therefore also investigate whether linear separability in the embedding space is present.

The success of most existing network embedding methods is due to the presence of strong homophily in many real networks~\citep{struc2vec}, i.e., the tendency of connected nodes to be similar.
Although node2vec relies primarily on homophily, it offers parameters that can emphasize structural equivalence in the embedding space to some extent. 
Structural equivalence differs substantially from homophily: two bridges may be considered structurally equivalent (or similar) despite not being connected and possibly far apart in the network.
Based on the classification performance in our experiments, we interpret these parameters to gain insight into which type of similarity may be more appropriate for road network analysis tasks.
We also evaluate the importance of other node2vec parameters.

Finally, the individual road categories and speed limits in our data set exhibits different degrees of homophily. We expect this to be problematic for subsequent classification on embeddings produced by a network embedding method.
Consequently, we investigate the relationship between homophily and classification performance on individual classes.

Our evaluation shows that given the right choice of classifier and parameters, node2vec can achieve high macro $F_1$ scores of $0.57$ for road category classification and $0.79$ for speed limit classification; improvements by a factor of $8.3$ and $11.5$, respectively, over choosing the most frequent class in the training set. In addition, our experiments suggest that additional hyperparameter tuning for both node2vec and the classifier can result in even better classification performance.
We also find that the class distribution in the embedding space reflects the class distribution in the road network, which, for the tasks we consider, results in a lack of linear separation in the embedding space.
By interpreting the classification performance for different values of node2vec parameters, we find that structural equivalence may be a more important type of similarity for road networks than homophily.
We also observed that a skew in the homophily of the classes results in a skew in classification performance on each class: highly homophilic classes achieve higher performances than less homophilic classes.

In summary, our contributions are as follows:
\begin{enumerate}
  \item We empirically evaluate node2vec on two road network analysis tasks and demonstrate that it can achieve $F_1$ scores that are up to $11.5$ times higher than choosing the most frequent class in the training set.
  \item We show how the geometric distribution of classes in the network is reflected in the embedding space, causing a loss of linear separability for the tasks in our case study.
  \item We demonstrate the impact of different node2vec parameters on classification performance, and show that emphasizing structural equivalence in the embedding space results in higher classification performance.
  \item We show that the reliance on neighborhood preservation results in a skew in classification performance towards favoring strongly homophilic classes.
\end{enumerate}

The rest of the paper is structured as follows.
In \cref{sec:related-work}, we discuss related methods that embed road network intersections and road segments and have been developed in parallel with network embedding methods.
In \cref{sec:network-embedding} we give the necessary background information on network embedding methods.  
In \cref{sec:experiments} we evaluate the node2vec embedding.
Finally, we represent our conclusions and discuss how we expect our findings to generalize to other network embedding methods and road network analysis tasks in \cref{sec:conclusion}.

\section{Related Work}\label{sec:related-work}
Embedding of road network intersections and segments is largely unexplored.
To the best of our knowledge, no previous work exists that investigates the application, adaptation, or extension of (general) network embedding methods to embed road network intersections and segments.
However, a few road network embedding methods---designed specifically for embedding road network intersections and segments---have been developed in parallel with (general) network embedding methods.
We proceed to review these methods.

Shahabi, et al.~\citep{Shahabi2003} use embeddings of intersections to improve the accuracy of $k$-nearest neighbor queries that find the $k$ nearest points of interest for moving vehicles.
Specifically, they embed intersections s.t.\ the shortest-path distance between two intersections is better approximated by the $L_{\inf}$ distance between the embedded intersections than by the Euclidean distance between their geographical points.
Liu et al.~\citep{Liu2011} further extend this approach to support private $k$-nearest neighbor queries, where the exact location of a vehicle is hidden.
As in our setting, only the road network structure is available, but we investigate whether network embedding methods may be applicable for a wider variety of road network tasks whereas \citep{Liu2011} aim to produce vectors that preserve a distance relationship between intersections for a specific task.


Road2Vec~\citep{road2vec} learns embeddings of road segments that capture traffic interactions. A traffic interaction between two road segments happens when the road segments co-occur within some distance of each other in a vehicle \gls{gps} trajectory. Using such co-occurrences, road2Vec assigns similar vector representations in the embedding space to road segments that interact frequently. Road2Vec requires a set of vehicle trajectories that covers all road segments in the road network to be able to learn meaningful similarities between road segments. This cannot be expected in general~\citep{yang2014using} and in our setting \gls{gps} trajectories are not available.

Another road segment embedding method, by \citet{realroad2vec}, relies on attribute information from nodes and edges in random walks to generate an embedding. Rather than capturing traffic interactions, this method captures a notion of structural equivalence between road segments based on the attributes of their surrounding road segments. They demonstrate that their method achieves superior performance on trip travel time prediction tasks compared to approaches based on traditional feature engineering with similar development effort~\citep{realroad2vec}.
This method and Road2Vec are quite similar to the network embedding method, node2vec, that we use in our case study. Like node2vec, these methods rely on samples of neighborhoods from the road network, either in the form of \gls{gps} trajectories or walks, and use techniques from natural language processing to produce the embeddings. 
However, contrary to our setting, this method assumes the presence of rich attribute information.

\section{Network Embedding}\label{sec:network-embedding}
We proceed to give the relevant background in network embedding.
We first give a general introduction to network embedding, and describe node2vec in detail.
We then briefly review other selected methods.

\subsection{A General Introduction} 
Let $G=(V, E)$ be the graph representation of a network. A network embedding is a function $\phi \colon V \cup E \xrightarrow{} \mathbb{R}^d$ that maps network elements to $d$-dimensional vectors. Network embedding methods often aim to learn a node embedding, but an edge may be embedded by aggregating the embeddings of the edge's incident nodes~\citep{node2vec}. For instance, an embedding of an edge $(v_1, v_2)$ can be found by concatenating the embeddings of its source and target nodes: $\phi(v_1, v_2) = [\phi(v_1)\phi(v_2)]$.

Deriving edge embeddings from node embeddings is useful for link prediction in social networks, where the task is to predict the existence of a missing edge~\citep{node2vec}. If all edges are known, an edge embedding can instead be produced by learning a node embedding for the dual graph representation of the network. In both cases, network embedding methods can be used to embed road segments.

Network embedding methods aim to map similar network elements to similar vector representations in the embedding space by optimizing an objective function that specifies the notion of similarity. In particular, network embedding methods find the node embedding $\phi$ that maximizes the following objective~\citep{deepwalk,node2vec,struc2vec,graphsage}:
\begin{equation}\label{eq:network-embedding-objective}
  \sum_{v \in V} \log \Big(\Pr(N(v) \mid \phi(v))\Big),
\end{equation}
where $N(v)$ is the neighborhood of node $v$ and $\Pr(N(v) \mid \phi(v))$ is the probability of observing the neighborhood $N(v)$ of $v$ given its feature representation $\phi(v)$. The probability $\Pr(N(v) \mid \phi(v))$ is computed using the softmax function~\citep{node2vec} or some approximation thereof. Other similar objectives have also been considered~\citep{cane,lane,sne,tridnr}.

The notion of neighborhood is not restricted to immediate neighbors, and many methods sample neighborhoods using random walks~\citep{deepwalk,node2vec, struc2vec, tridnr}. In such cases, the notion of node neighborhood is defined in terms of a set of walks across the node s.t.\, given a walk $(v_1, \dots, v_n)$, the neighborhood of a node $v_i$ is the $c$ preceding and succeeding nodes in the walk $\{v_{i-c}, \dots, v_{i-1}, v_{i+1}, \dots, v_{i+c}\}$, where $c$ is the context size. Neighborhood sampling using random walks can scale to very high-degree networks; examples include social networks where nodes have thousands or even millions of neighbors\footnote{The YouTube account on Twitter has $70.6$ million followers at the time of writing.}.

Maximizing \cref{eq:network-embedding-objective} results in an embedding that is optimized to preserve node neighborhoods (according to $N(v)$) in the embedding space~\citep{node2vec,struc2vec}. Intuitively, \cref{eq:network-embedding-objective} suggests that the similarity between two nodes $u$ and $v$ in the embedding space is proportional to the size the overlap of their neighborhoods. For such an embedding to be useful, the underlying network must exhibit \emph{homophily}: the tendency of network elements to be connected to similar network elements~\citep{struc2vec}. From the perspective of a supervised learning task, homophily means that neighboring nodes are likely to have the same label. An embedding that preserves node neighborhoods places neighboring nodes close in the embedding space and therefore produces useful feature vectors for subsequent machine learning under the homophily assumption. 

We anticipate that producing a good embedding for road networks while relying on neighborhood preservation is difficult for three reasons. First, a road network is a spatial construct and inaccuracies in the spatial representation of a road network may result in false or missing edges in the graph representation.
Next, countries such as Denmark have islands that may result in disconnected subgraphs. The effect of such subgraphs on the training of the network embedding is not immediately obvious, but it may add noise to the training procedure, resulting in a reduced degree of neighborhood preservation. Finally, island nodes and main land nodes are not similar according to \cref{eq:network-embedding-objective} since they are neither neighbors nor share neighbors. 

\subsection{node2vec}\label{sec:node2vec}
We use the embedding method node2vec in our case study. We first discuss node2vec's training objective and then proceed to look at the flexible sampling strategy employed by node2vec.

First, node2vec optimizes \cref{eq:network-embedding-objective} directly and uses random walks to represent node neighborhoods. For each node $v$ in a graph, node2vec samples $r$ walks of maximum length $l$ starting from $v$ s.t.\ the next node visited in the walk is chosen at random among the neighbors of the last node in the walk. Therefore, the neighborhood of a node $v$ is distributed over the walks. To illustrate this, we rewrite \cref{eq:network-embedding-objective} to reflect this distributed neighborhood in \cref{eq:node2vec-objective}. Note that \cref{eq:node2vec-objective} assumes that the neighborhoods of a node $v$ w.r.t.\ each walk are conditionally independent given $\phi(v_i)$.

\begin{equation}\label{eq:node2vec-objective}
  \sum_{W \in \mathcal{W}} \sum_{v_i \in W} \log \Big(\Pr\big( N^W(v_i) \mid \phi(v_i)\big)\Big)
\end{equation}
Here, $\mathcal{W}$ is a set of $r\cdot |V|$ walks $W=\{v_1, \dots, v_k\}$ s.t.\ $k \leq l$ and $N^W(v_i) = \{v_{i-c}, \dots, v_{i-1}, v_{i+1}, \dots, v_{i+c} \}$ is the neighbors of node $v_i$ with respect to walk $W$. The context size $c$ adjusts the number of preceding and succeeding nodes in the walk to consider the neighbors of $v_i$. In general, $c$ should be selected s.t.\ $2c << l$ to avoid frequently padding $N^W(v_i)$ with ''null'' nodes.


Next, node2vec samples $w$ walks with a maximum length of $l$ using a second-order biased random walk, where $w$ and $l$ are hyperparameters of node2vec. Given that the random walk has just traversed an edge $(v_{i-1}, v_{i})$ in an unweighted graph and now resides at node $v_i$, the probability of visiting a node $v_{i+1}$ is~\citep{node2vec}:
\begin{equation}\label{eq:biased-random-walk}
  \Pr(v_{i+1} \mid v_{i}, v_{i-1}) = \frac{1}{Z} \cdot \begin{cases}
    \frac{1}{p} & \text{if $d(v_{i-1}, v_{i+1}) = 0$}  \\
    1 & \text{if $d(v_{i-1}, v_{i+1}) = 1$}  \\
    \frac{1}{q} & \text{if $d(v_{i-1}, v_{i+1}) = 2$}  \\
    0 & \text{if $(v_{i}, v_{i+1}) \notin E$}
  \end{cases},
\end{equation}
Here, $Z$ is a normalization constant and $d(u, v)$ returns the distance between two nodes $u$ and $v$. $p$ and $q$ are hyperparameters that change the behavior of the walk to behave as a \gls{bfs}, \gls{dfs}, or something in-between. A \gls{bfs} emphasizes homophily and a \gls{dfs} emphasizes structural equivalence as the type of similarity to capture in the embeddings~\citep{node2vec}.

The \emph{return parameter} $p$ adjusts the probability of revisiting the previous node $v_{i-1}$ in the walk and restricts the number of different nodes visited in the search. Low values of $q$ is equivalent to restricting the search depth in a \gls{dfs} and equivalent to restricting the number of neighbors to explore in a \gls{bfs}.
The \emph{in-out parameter} $q$ adjusts the probability of visiting different neighbor of $v_{i-1}$. In effect, $q$ adjust the behavior of the walk to become more \gls{bfs}-like at high values and more \gls{dfs}-like at low values.

Although node2vec can capture structural equivalence to some extent, even using an actual \gls{dfs} to sample neighborhoods results in neighborhoods that can include nodes up to a maximum distance of $c$.
Thus, the structural equivalence emphasized by the walks is local to the area of the network from which it is sampled.
This makes the walks unable to capture structural equivalences between, e.g., bridges that are far apart in the network, and thus node2vec still primarily relies on homophily in networks with large diameters such as country-sized road networks.

The embedding function $\phi$ is generated using a single-layer neural network. A node $v_i$ is embedded in the $d$ hidden units of the (hidden) embedding layer.
The resulting $d$-dimensional embedding is then used to predict $v_i$'s neighbors (w.r.t.\ a walk) at the output layer by using the softmax function (or some approximation thereof) to compute the probability $\Pr( N^W(v_i) \mid \phi(v_i))$ in \cref{eq:node2vec-objective}.

\subsection{Other Approaches}
The discussion of network embedding methods has thus far been relatively focused. For completeness, we briefly review other approaches to network embedding; however, we note that all of the following methods to some extent preserves node neighborhoods as in \cref{eq:network-embedding-objective}. Fundamentally, network embedding methods differ primarily in either the choice of neighborhood function or in how (and if) they incorporate node or edge attributes in the embedding.

\subsubsection{Alternative Neighborhood Function}
So far, a node neighborhood has been expressed as all nodes within $c$ hops of a node or some subset thereof. As discussed previously, this leads to a neighborhood preserving embedding. The \emph{struc2vec} embedding method~\citep{struc2vec} changes the neighborhood sampling procedure by sampling walks from a multi-layered graph. A node $v$ in a layer $k$, is connected by an edge to all nodes at exactly distance $k$; hence each layer represents the neighborhood of $v$ at different ''zoom'' levels. The walk can choose to stay at the current layer or to proceed to a higher layer depending on a heuristic designed to preserve structural similarity. This enables struc2vec to effectively skip nodes in the walk.
The worst-case time and space complexities of struc2vec are $\mathcal{O}(|V|^3)$ and $\mathcal{O}(|V|^2)$, respectively.
These complexities can be improved significantly through various optimization methods, but continues to be super-linear~\citep{struc2vec}, which hinders its applicability to large road networks.

\subsubsection{Incorporation of Attributes}
Network embedding methods that incorporate attributes use them as input to the embedding method~\citep{sne,graphsage} and possibly include a sub-objective in the objective function that encourages nodes or edges with similar attributes to obtain similar vector representations in the embedding space~\citep{tridnr,lane,cane}. This allows such approaches to compute a notion of similarity even between disconnected or distant nodes. However, these methods still aim to preserve node neighborhoods in the embedding space to some extent.

\section{Experimental Study}\label{sec:experiments}
To investigate the suitability of network embedding methods for road network analysis, we evaluate the node2vec method on two road segment classification tasks and report the results.

The existing embedding literature commonly uses linear classifiers to emphasize the quality of the embedding over the complexity of the classifier, and typically such classifiers achieve high classification performance.
This suggest that network embedding methods tend to make the classification problems in the literature linearly separable in the embedding space.
Given that such methods are typically applied to, e.g., social networks, that exhibit characteristics different from road networks, it is not clear whether linear separability in the embedding space extends to road networks.
We therefore investigate whether this property is present for the classification tasks that we consider.

Network analysis tasks typically depend on homophily or structural equivalence as the notion of similarity~\citep{node2vec}.
To investigate which type of similarity is appropriate for road network analysis tasks, we exploit the interpretability of the node2vec random walk parameters $p$ and $q$, as discussed in \cref{sec:node2vec}, by investigating the impact of these parameters on classification performance.
We also investigate the impact of the node2vec architecture parameters, the dimensionality $d$ and context size $c$, to gain insight into how changes in the node2vec architecture influences classification performance. 

The success of most existing network embedding methods is due to the presence of strong homophily in many real networks~\citep{struc2vec}. In our data set, the individual classes exhibit different degrees of homophily, which suggest that this is problematic for network embeddings methods.
We therefore investigate the relationship between homophily and classification performance on individual classes.

\subsection{Data Set}
We have extracted the spatial representation of the Danish road network from \gls{osm}~\citep{osm}. We represent the road network as a directed multigraph, where each node represents an intersection or the end of a road and each edge $(u, v)$ represents a road segment that enables travel from node $u$ to node $v$. This yields a graph consisting of $583,816$ nodes and $1,291,168$ edges.

The data from \gls{osm} contains two road segment attributes, road category and speed limit.
We further augment the data set with additional speed limit information from the Danish municipalities of Aalborg and Copenhagen.
This results in $1,291,168$ road segments (i.e., all edges) labeled with one of $9$ road categories
and $163,043$ road segments (\texttildelow$13\%$ of all edges) labeled with one of $10$ speed limits. 
We also note that the speed limits are not distributed evenly geographically: speed limits in major cities are over-represented in the data.

As discussed in \cref{sec:network-embedding}, network embedding methods are suited for networks that are homophilic w.r.t.\ the attributes of interest. We therefore measure the homophily for both the road category and speed limit attributes.

We measure the homophily of an attribute value $a$ in a directed network $G = (V, E)$ as \emph{the empirical probability that an edge $e_1 = (u, v)$ is adjacent to an edge $e_2 = (v, w)$ that has attribute value $A(e_2) = a$ given that $e_1$ has attribute value $A(e_1) = a$}, i.e.,
\begin{flalign*}
  H_G^c &= \Pr(\mathit{A}(v, w) = a \mid \mathit{A}(u, v) = a) \\
        &= \sum_{(u, v) \in E} \sum_{(v, w) \in E} \frac{\mathbbm{1}[\mathit{A}(u, v) = \mathit{A}(v, w)]}{Z},
\end{flalign*}
where $A(v_1, v_2)$ is the attribute value of an edge $(v_1, v_2)$ and $Z$ is a normalization constant.

We compute the homophily of network $G$ with regards to an attribute $A = \{a_1, \dots, a_m\}$ as follows.
\begin{equation*}   
  H_G^A = \sum_{a \in A} \frac{H_G^a}{|A|}
\end{equation*}
We summarize the data set statistics in \cref{tab:data-set-statistics}. As can be seen, there is a homophily of $H_G^L=72.3\%$ for road categories and $H_G^L = 75.8\%$ for speed limits. A notable outlier is the road category ''Motorway Approach/Exit'' which has a homophily of $36.8\%$.

\begin{table}
  \centering
  \caption{Data Set Statistics for Road Categories and Speed Limits\label{tab:data-set-statistics}}
  \scriptsize
  \begin{tabular}{rlcr}
    \toprule
    & \emph{Class} & \emph{Homophily} & \emph{Frequency} \\
    \midrule
    \multicolumn{4}{l}{\emph{Road Categories}} \\
          & Residential  & $90.4\%$ & \textbf{$\mathbf{570,820}$ ($\mathbf{44.2\%}$)} \\
          & Service      & $72.7\%$ & $278,985$ ($21.6\%$) \\
          & Unclassified & $78.0\%$ & $257,726$ ($20.0\%$)\\
          & Tertiary     & $70.2\%$ & $103.830$ ($8.04\%$) \\
          & Secondary    & $70.6\%$ & $52,021$  ($4.03\%$)\\
          & Primary      & $71.7\%$ & $22,255$  ($1.72\%$)\\
          & Motorway     & $78.2\%$ & $2,236$ ($0.173\%$) \\
          & Motorway Approach/Exit  & $\mathbf{36.8\%}$ & $1,749$ ($0.135\%$)  \\
          & Trunk        & $81.7\%$ & $1,546$ ($0.120\%$)  \\
    \emph{Mean/Total} &  & $72.3\%$ & $1,291,168$ ($100\%$) \hspace{4.5pt} \\
    \midrule
    \multicolumn{4}{l}{\emph{Speed Limits}} \\
          & $50$ & $82.2\%$ & \textbf{$\mathbf{85,377}$ ($\mathbf{52.4\%}$)} \\
          & $80$ & $73.1\%$ & $37,750$ ($23.2\%$) \\
          & $40$ & $79.7\%$ & $11,830$ ($7.26\%$) \\
          & $60$ & $64.9\%$ & $10,112$ ($6.20\%$) \\
          & $30$ & $78.4\%$ & $9,093$  ($5.58\%$) \\
          & $70$ & $63.2\%$ & $4,481$  ($2.75\%$) \\
          & $20$ & $80.7\%$ & $1,383$ ($0.848\%$) \\
          & $110$ & $70.5\%$ & $1,103$ ($0.677\%$) \\
          & $90$ & $72.9\%$ & $1,087$ ($0.664\%$) \\
          & $130$ & $71.5\%$ & $827$ ($0.507\%$) \\
    \emph{Mean/Total} &   & $75.8\%$ & $163,043$ ($100\%$) \\
    \bottomrule
  \end{tabular}
\end{table}


\subsection{Experiment Design}
We use the node2vec~\citep{node2vec} embedding to classify road categories and speed limits as follows.
\begin{enumerate}
  \item We first sample $r=10$ walks, starting from each intersection in the road network, with a maximum length of $l=80$ using the biased random walk in \cref{eq:biased-random-walk} parameterized by the return parameter $p$ and the in-out parameter $q$. 
  \item We then learn the embedding using these walks as input.
  \item Finally, we train a classifier for each of the road segments classification tasks using these embeddings.
\end{enumerate}
To investigate the linear separability in the embedding space, we consider both a linear and a non-linear classifier:
\begin{itemize}
  \item a one-vs-rest logistic regression model as in the original node2vec paper~\citep{node2vec}, and
  \item a random forest classifier~\citep{Breiman1996}, a powerful ensemble model, with $10$ decision trees.
 \end{itemize}

After learning the embedding, we generate a feature vector for each road segment by concatenating the embeddings of its source and target nodes as discussed in \cref{sec:network-embedding}.
For both road category and speed limit classification, we randomly choose $50\%$ of the labels without replacement for training and use the remaining $50\%$ for testing.
To deal with the large class imbalance in our data set (see \cref{tab:data-set-statistics}), we randomly over-sample all classes with replacement s.t.\ the frequency of all classes in the training set match the frequency of the majority class.
We then train road category and speed limit classifiers using these feature vectors to represent each labeled road segment in the over-sampled training set.
Finally, we evaluate the classifier on the two road segment classification tasks using the macro $F_1$ score, which is commonly used in the network embedding literature~\citep{deepwalk,node2vec, tridnr, graphsage}.
The macro $F_1$ scores punishes poor performance equally across all classes, rather than rewarding good performance on the very frequent classes in our imbalanced data set.

To find the best node2vec parameter configuration for each of the classifiers, we learn node2vec embeddings with different configurations of return parameter $p$, the in-out parameter $q$, the context size $c$, and the dimensionality $d$.
We explore configurations based on the following possible parameter values: $p \in \{0.25, 0.5, 1, 2, 4\}$, $q \in \{0.25, 0.5, 1, 2, 4\}$, $c \in \{1, 5, 10, 15, 20, 25, 30 \}$, and $d \in \{64, 128, 256\}$.
We use the baseline values of return parameter $p=1$, in-out parameter $q=1$, and context size $c=10$. We then explore different values for $p$, $q$, and $c$ at different values of $d$ while keeping the other two parameters at their baseline values.
Although use of these parameter configurations does not result in an exhaustive search of the parameter space, they do allow us to evaluate the impact of each parameter on classification performance.
Finally, the $c$ and $d$ parameters do not influence the walk sampling procedure, and we therefore reuse walks for parameter configurations that have the same $p$ and $q$ values.

To establish baseline performances for evaluation, we use two simple classifiers that classify road segments using only the statistics of the training set:
\begin{itemize}
  \item \emph{Most Frequent}: Always predicts the most frequent class in the training set.
  \item \emph{Empirical Sampling}: Draws a class at random from the empirical distribution of classes in the training set.
\end{itemize}

\COMMENT{
\subsection{Experimental Setting}
\todo{Update}
We perform our experiments using \emph{Python 3.6}. For in-memory graph representations, we use \emph{networkx}~\citep{networkx} and \emph{igraph}~\citep{igraph}\footnote{Python bindings available through the \emph{python-igraph} package. }.\@ We use igraph solely for shortest path computations when computing geodesic distances and shortest route distances in \cref{sec:distance-preservation}.

We use the reference implementation of \emph{node2vec}~\citep{node2vec}\footnote{Implementation of node2vec is available at \url{https://github.com/aditya-grover/node2vec}.} to train the DeepWalk embedding by setting the node2vec parameters $p=1$ and $q=1$. The node2vec implementation generates walks and feeds them to the \emph{gensim}~\citep{gensim} implementation of \emph{word2vec}~\citep{word2vec} to generate a node embedding. We use the scikit-learn~\citep{scikit-learn} implementation of one-vs-rest logistic regression for the edge classification tasks and imbalanced-learn~\citep{imbalanced-learn} for oversampling minority labels.
}

\subsection{Road Segment Classification}
\begin{figure}
  \begin{subfigure}{\columnwidth}
    \includegraphics[width=\columnwidth, trim={0 0.7cm 0 0.475cm}, clip]{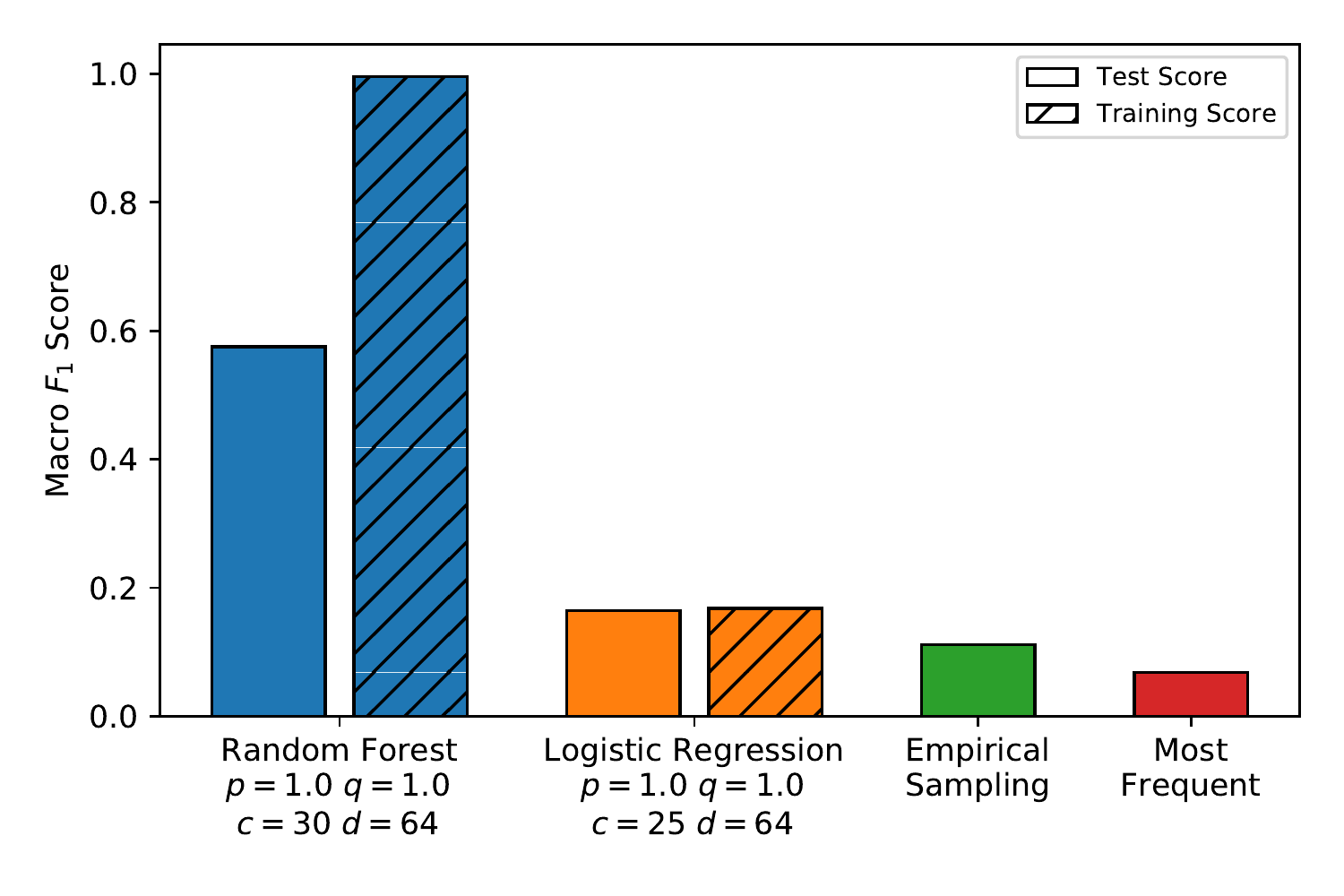}
    \caption{Road Category Classification.\label{fig:summary-category}}
  \end{subfigure}
  \begin{subfigure}{\columnwidth}
    \includegraphics[width=\columnwidth, trim={0 0.7cm 0 0.475cm}, clip]{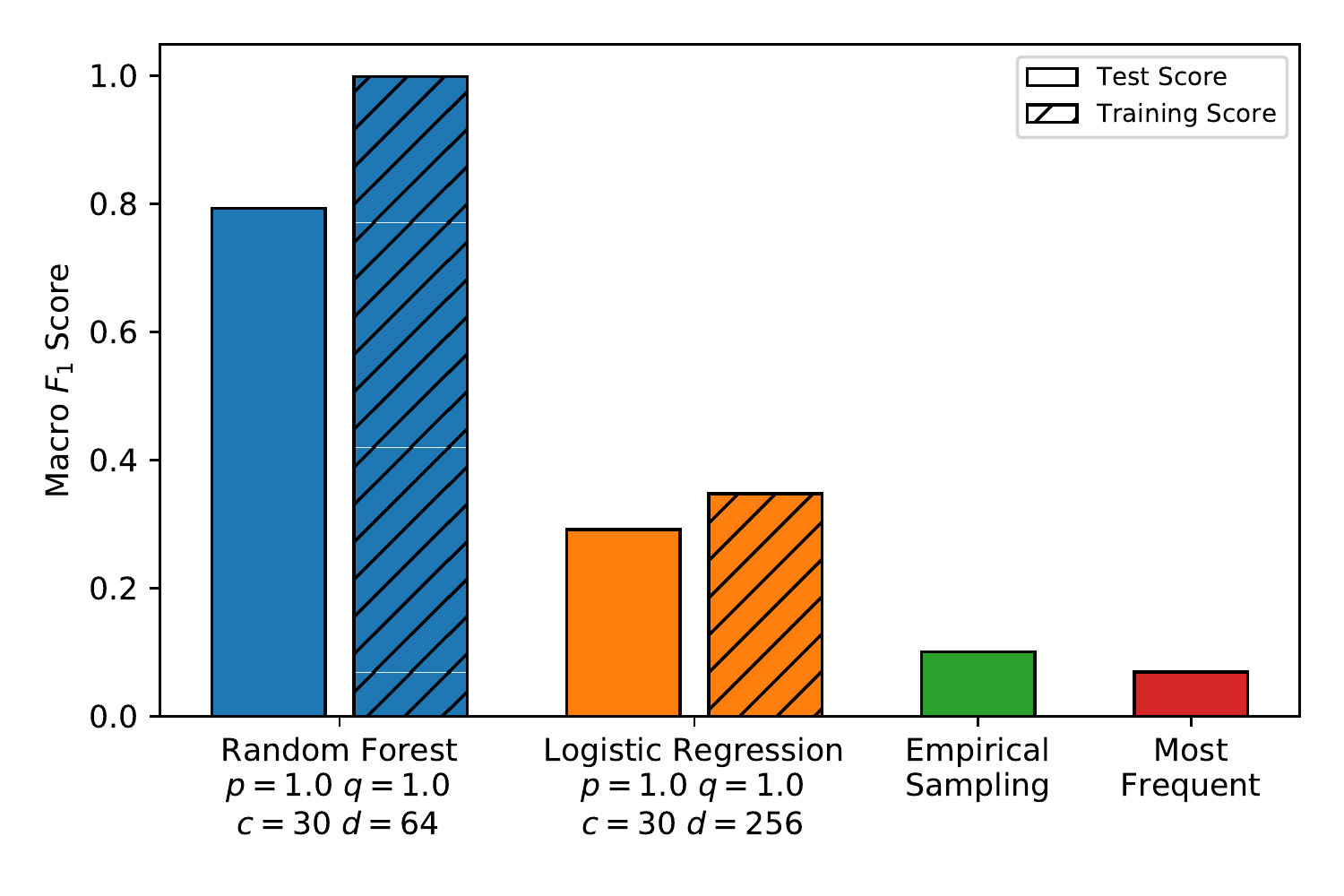}
    \caption{Speed Limit Classification.}
  \end{subfigure}
  \caption{Macro $F_1$ scores on the training and test sets for the best performing embedding for each classifier with baselines for comparison on (a) road category classification and (b) speed limit classification.\label{fig:summary}}
\end{figure}

We show the performance of the best performing node2vec parameter configuration for road category and speed limit classification using logistic regression and random forests (with baselines for comparison) in \cref{fig:summary}.
As can be seen, the choice of classifier has a large impact on classification performance.
The macro $F_1$ score using a random forest is $0.57$ for road category classification and $0.79$ speed limit classification. This is roughly three times higher than the score achieved using logistic regression, and $8.3$ and $11.5$ times higher than the two baselines, respectively.
We expect that these results can be improved by choosing appropriate $p$ and $q$ parameter values, as we shall discuss in \cref{sec:homophily-vs-structural-equivalence}.
In addition, the random forest achieves a near-perfect score on the training set for both tasks, which suggests that it is overfitting and could conceivably achieve even higher performance by tuning its hyperparameters.
On the other hand, the logistic regression model achieves, roughly equal performance on the training and test sets for both classification tasks, and is only marginally better than the Empirical Sampling baseline on road category classification, as shown in \cref{fig:summary-category}.

\subsection{Linear Separability}
It is not surprising that the random forest model outperforms the logistic regression model given that the random forest uses advanced ensemble learning techniques, such as boosting. 
However, the low score on both training and test sets shown in \cref{fig:summary} suggests that the logistic regression model is both unable to fit the training data and generalize to the test data.
This suggests that the logistic regression model is unable to find good linear decision boundaries and thus that the classes are not linearly separable in the embedding space.

\begin{figure}[t]
  \centering
  \begin{subfigure}{\columnwidth}
    \centering
   \includegraphics[width=0.995\columnwidth, trim={0 0.75cm 0 1.1cm}, clip]{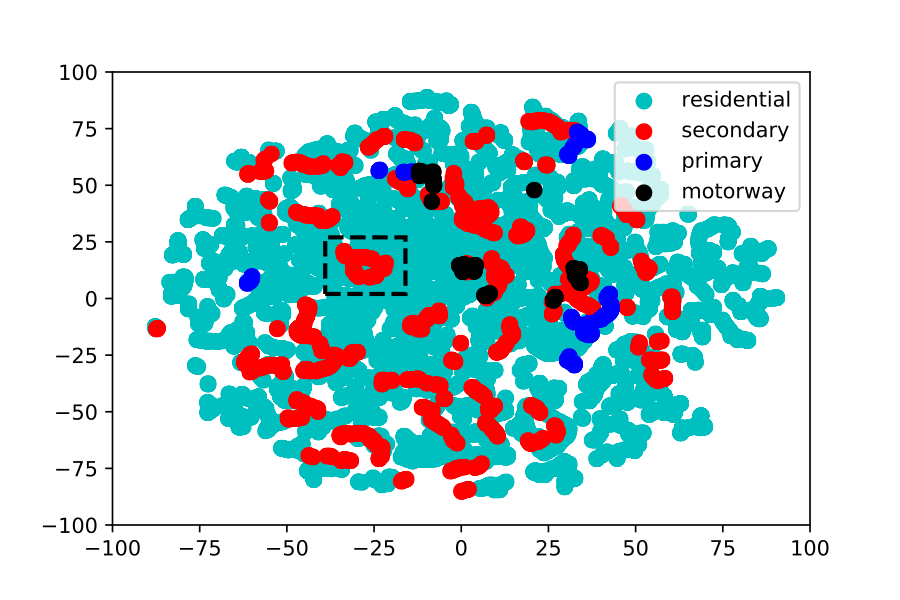}
    \caption{Embedded road segments.\label{fig:tsne-clusters-embedding}}
  \end{subfigure}
  \begin{subfigure}{\columnwidth}
    \centering
      \includegraphics[width=0.8375\columnwidth, height=4.65cm]{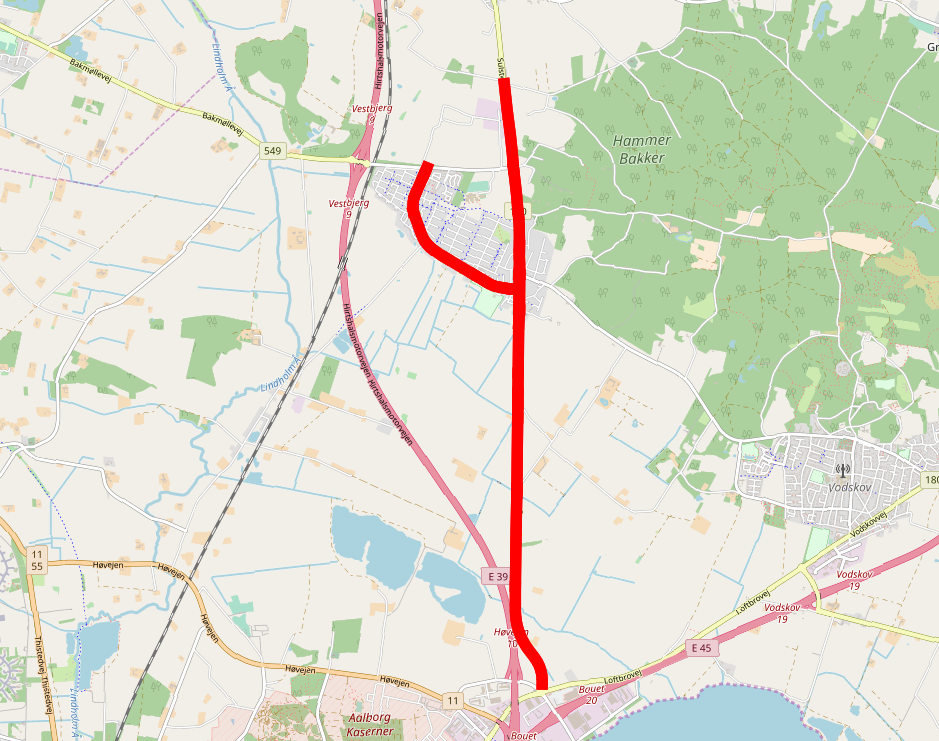}
    \caption{Spatial representations of the framed cluster in \cref{fig:tsne-clusters-embedding}.\label{fig:tsne-clusters-spatial}}
  \end{subfigure}
  \caption{Two-dimensional \gls{tsne} projections of the road segment embeddings in Aalborg Municipality.\label{fig:tsne-clusters}}
\end{figure}

We investigate the linear separability in the embedding space in detail by embedding the road network of Aalborg Municipality, Denmark, using node2vec with baseline parameters and visualizing the position of the road segments in the embedding space by reducing the dimensionality of their feature vectors.
We use Barnes-Hut \gls{tsne}~\citep{tsne} to project the feature vectors into two dimensions and plot them in \cref{fig:tsne-clusters}.
Each point in the plot represents a directed road segment, and is colored according to its road category.
For illustrative purposes, we cover only four of the road categories in the plot.

As can be seen in \cref{fig:tsne-clusters-embedding}, road categories are not well-separated in the two-dimensional \gls{tsne} projection.
The residential road segments are scattered almost uniformly over the embedding space, and the remaining categories are scattered in several smaller clusters.
Each such cluster corresponds to a homophilic neighborhood (or area) in the road network.
For instance, the cluster of secondary road segments highlighted in \cref{fig:tsne-clusters-embedding} corresponds to the $57$ road segments highlighted in red in \cref{fig:tsne-clusters-spatial}.
These scattered clusters suggest that it is difficult to find good linear decision boundaries for the logistic regression model, which is further supported by its poor performance on both the training and test sets as shown in \cref{fig:summary}.

The lack of separation in the embedding space reflects the lack of separation in the road networks due to the neighborhood preserving properties of the embedding method: for example, several clusters of secondary segments can be found in different areas of the road network. 
Although we have focused the discussion on road categories, speed limits show a similar data distribution.
Given that logistic regression is unable to find linear decision boundaries in the embedding space that are competitive with the random forest classifier, we focus on the random forest classifiers for our experimental results in the remainder of the paper.

\subsection{Homophily or Structural Equivalence}\label{sec:homophily-vs-structural-equivalence}
Next, we examine how the return parameter $p$ and the in-out parameter $q$ influences classification performance.

Recall from \cref{sec:node2vec} that low values of $p$ increases the probability of revisiting the previous node in the walk and that the parameter $q$ makes the walk behave like a \gls{dfs} for low values of $q$ and like a \gls{bfs} for high values of $q$.
Intuitively, $p$ controls the breadth or depth of the search in terms of how many different nodes are visited, while $q$ controls to which extent the random walk behaves more like a \gls{bfs} or a \gls{dfs}.
In addition, recall that a \gls{bfs}-like walk emphasizes homophily in the embedding space and a \gls{dfs}-like walk emphasizes structural equivalence (but restricted by the neighborhood defined by the walk).
Thus, this structural equivalence is local, in the sense that structural equivalence between, e.g., bridges in different parts of the country cannot be captured in networks with large network diameter. However, node2vec may still provide insight into the appropriate type of similarity for the road segment classification tasks.

\begin{figure}[h]
  \begin{subfigure}{\columnwidth}
    \includegraphics[width=\columnwidth, trim={0 0.525cm 0 0.425cm}, clip]{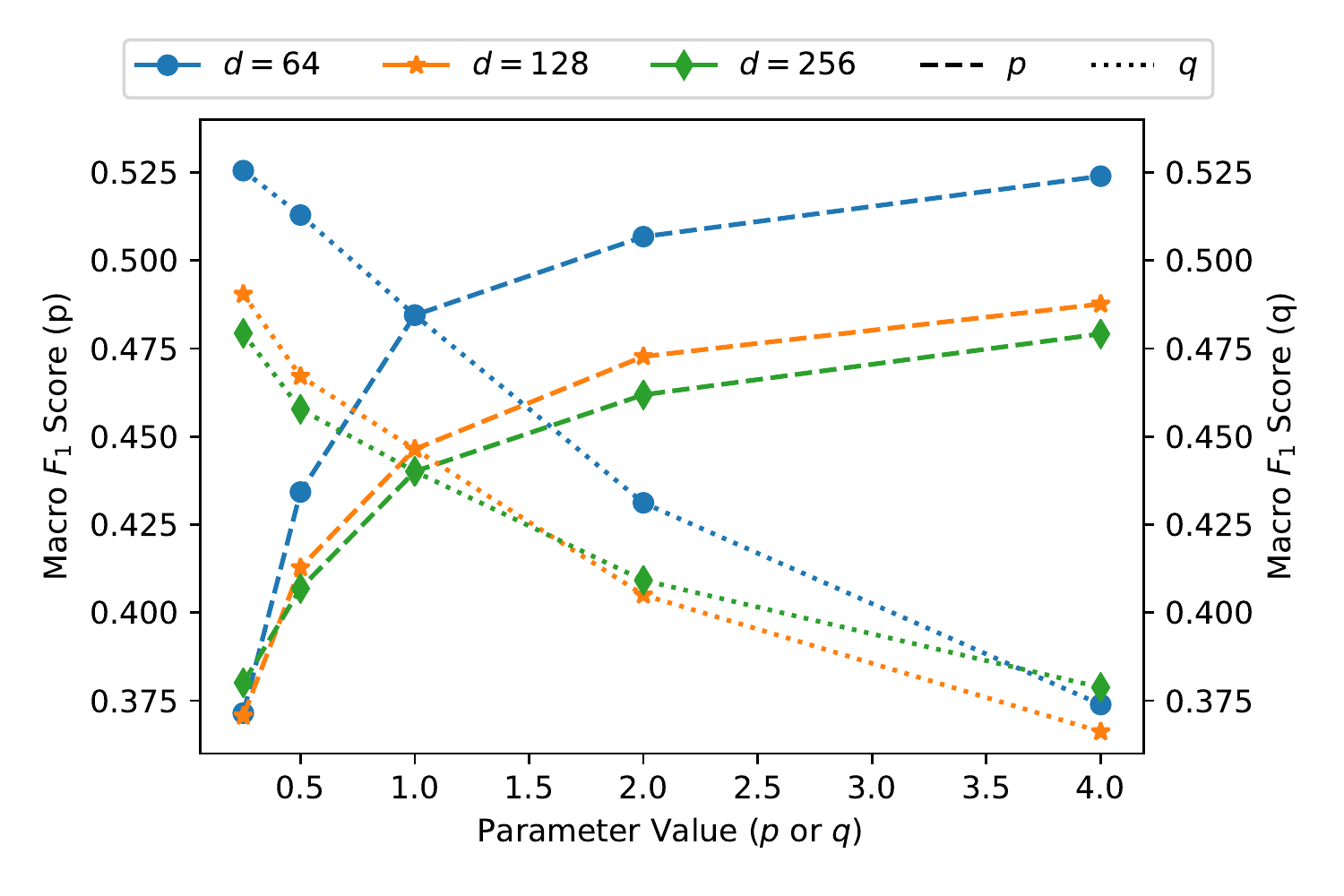}
    \caption{Road Category Classification.}
  \end{subfigure}
  \begin{subfigure}{\columnwidth}
    \includegraphics[width=\columnwidth, trim={0 0.525cm 0 0.425cm}, clip]{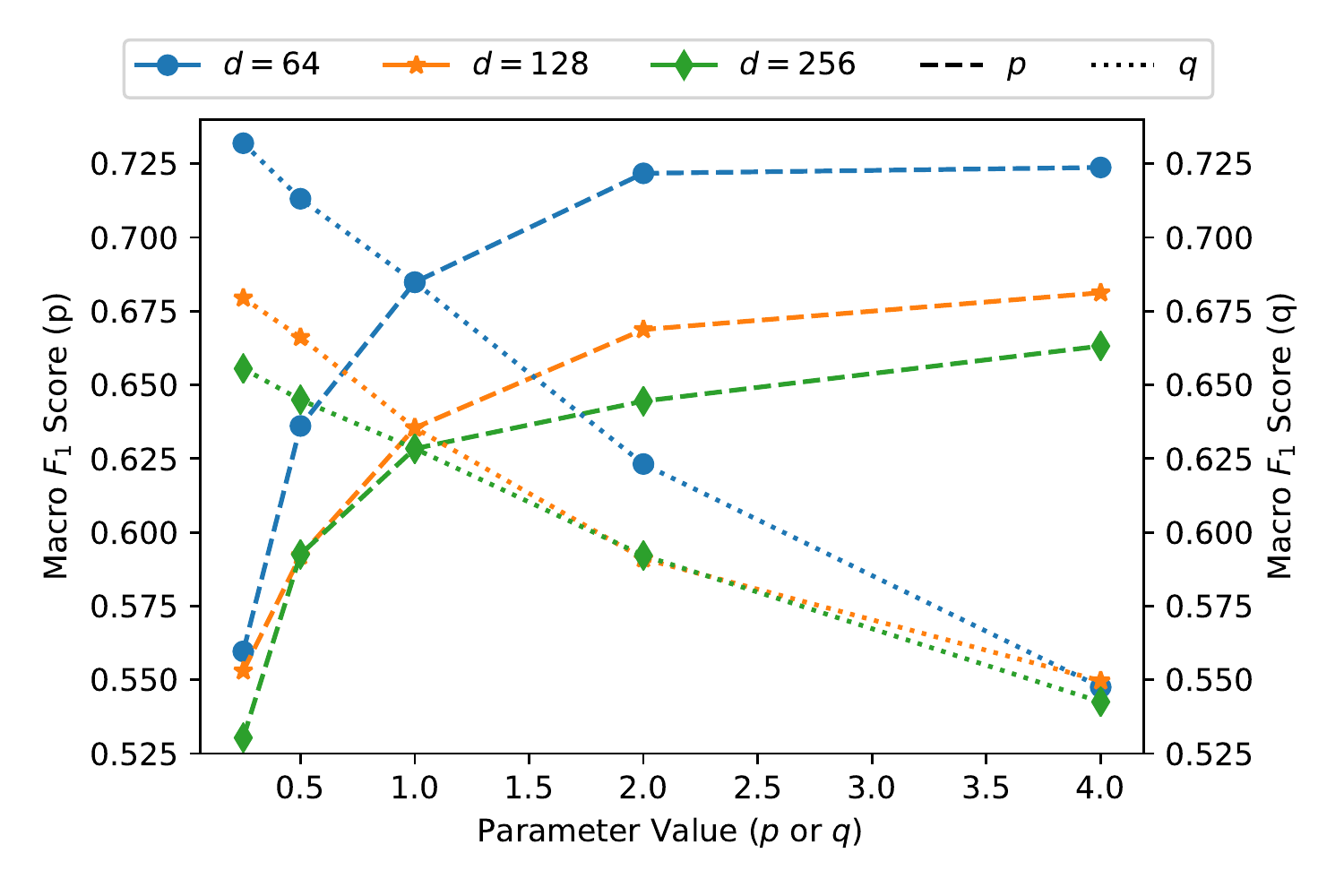}
    \caption{Speed Limit Classification.}
  \end{subfigure}
    \caption{The effects of $p$ (dashed) and $q$ (dotted) on (a) road category classification and (b) speed limit classification using a random forest classifier.\label{fig:p-q-performance}}
\end{figure}

We plot the classification performance for the random forest classifier for different values of $p$ and $q$ on the two road segment classification tasks in \cref{fig:p-q-performance}.
As shown in the figure, classification performance is highest at high values of $p$ and low values of $q$, regardless of the dimensionality.
The high performance at high values of $p$ suggests that exploring more nodes in the walk sampling procedure is beneficial.
The high performance at low values of $q$ suggests that a \gls{dfs}-like neighborhood exploration is superior to a \gls{bfs}-like neighborhood exploration.
We do not reach a saturation point for the $q$ values which suggest that the more \gls{dfs}-like the walk, the better.
Thus, our results suggest that structural equivalence should be emphasized over homophily in the embedding space. 

\begin{figure}[h]
    \includegraphics[width=\columnwidth, trim={0 0.55cm 0 0.48cm}, clip]{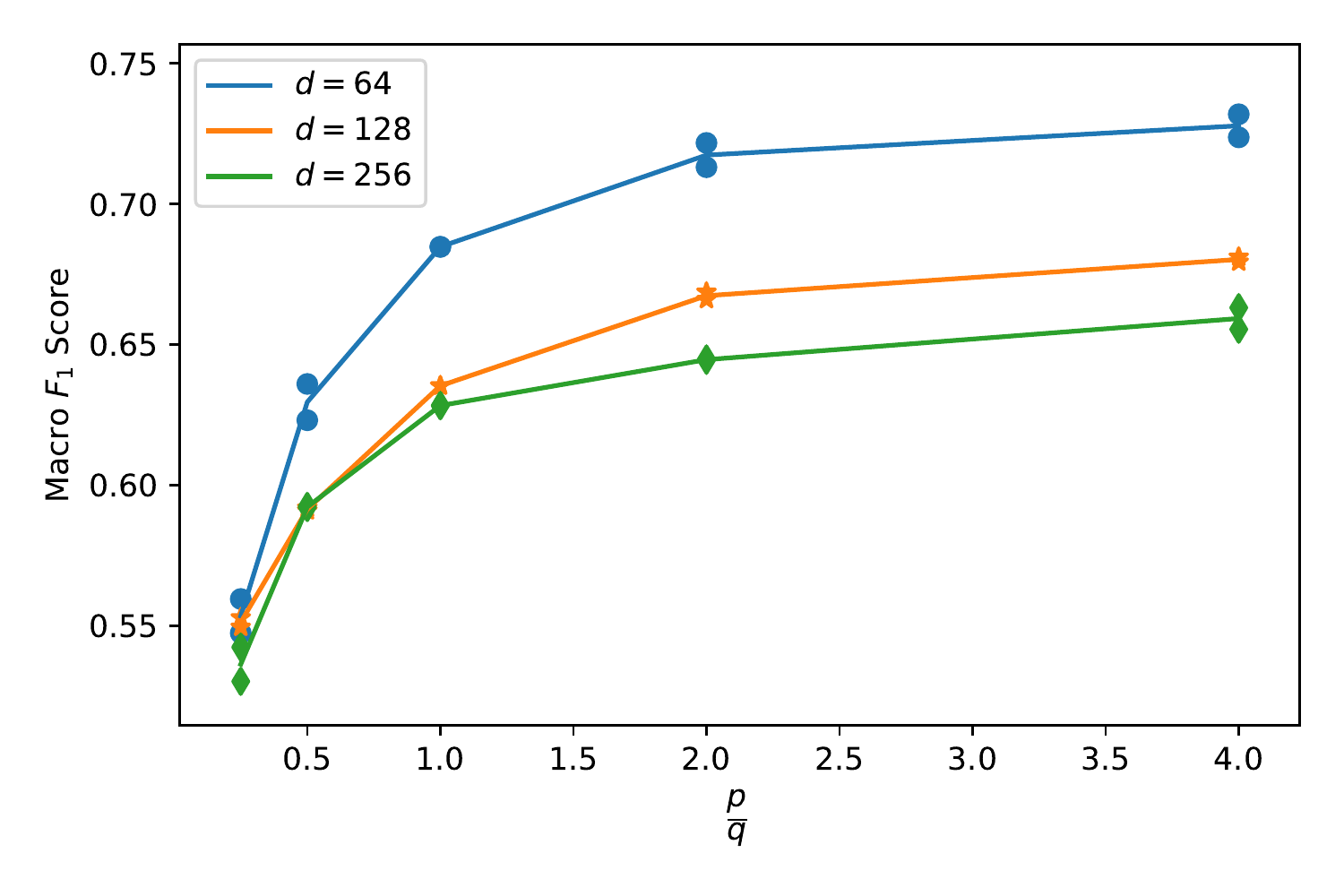}
  \caption{The relationship between $\frac{p}{q}$ and classification performance on the speed limit classification task using a random forest classifier.\label{fig:p-q-ratio}}
  \vspace*{-0.35cm}
\end{figure}

We make the additional observation that $p$ and $q$ adversely affect each other in \cref{eq:biased-random-walk}: a high $p$ value reduces the probability of visiting a node at distance two from the previously visited node and a high $q$ value reduces the probability of returning to a previously visited node.
We therefore propose that the ratio between these parameters is more important than their absolute values.

For the parameter configurations we explore in our experiments, a ratio of, e.g., $\frac{p}{q} = 4$ can occur for values $p=1$ and $q=0.25$ or values $p=4$ and $q=1$.
We therefore plot the data points for both the case where $p > q$ and the case where $q > p$ by taking the mean of their associated macro $F_1$ scores; see \cref{fig:p-q-ratio}.
We also investigated whether $p > q$ should be preferred over $q > p$ for each ratio, but we found no trend for the values that we examined to suggest that one combination of $p$ and $q$ values should be preferred over the other given that they have the same $\frac{p}{q}$ ratio.

As shown by the figure, a larger $\frac{p}{q}$ ratio results in a higher classification performance on the speed limit classification task.
We observe the same pattern on road category classification.
This suggests that emphasizing structural equivalence in the embedding space is more important than emphasizing homophily for the tasks we consider which is consistent with our previous observations on the values of $p$ and $q$.

\subsection{Architectural Parameters}
We also investigate the impact of context size $c$ and dimensionality $d$ on classification performance.
These parameters adjust the amount of neighborhood information available and the number of weights available for node2vec.

The context size $c$ and the return parameter $p$ serve somewhat similar roles since they both control the number of different nodes included in the neighborhood: large values of $c$ include more nodes from the walks in the node neighborhood whereas small values of $p$ makes it less likely that we revisit nodes and thus increases the number of different nodes that occur in the neighborhood of a node. We therefore expect that larger context sizes results in higher performance.

\begin{figure}
    \includegraphics[width=\columnwidth, trim={0 0.275cm 0 1.2cm}, clip]{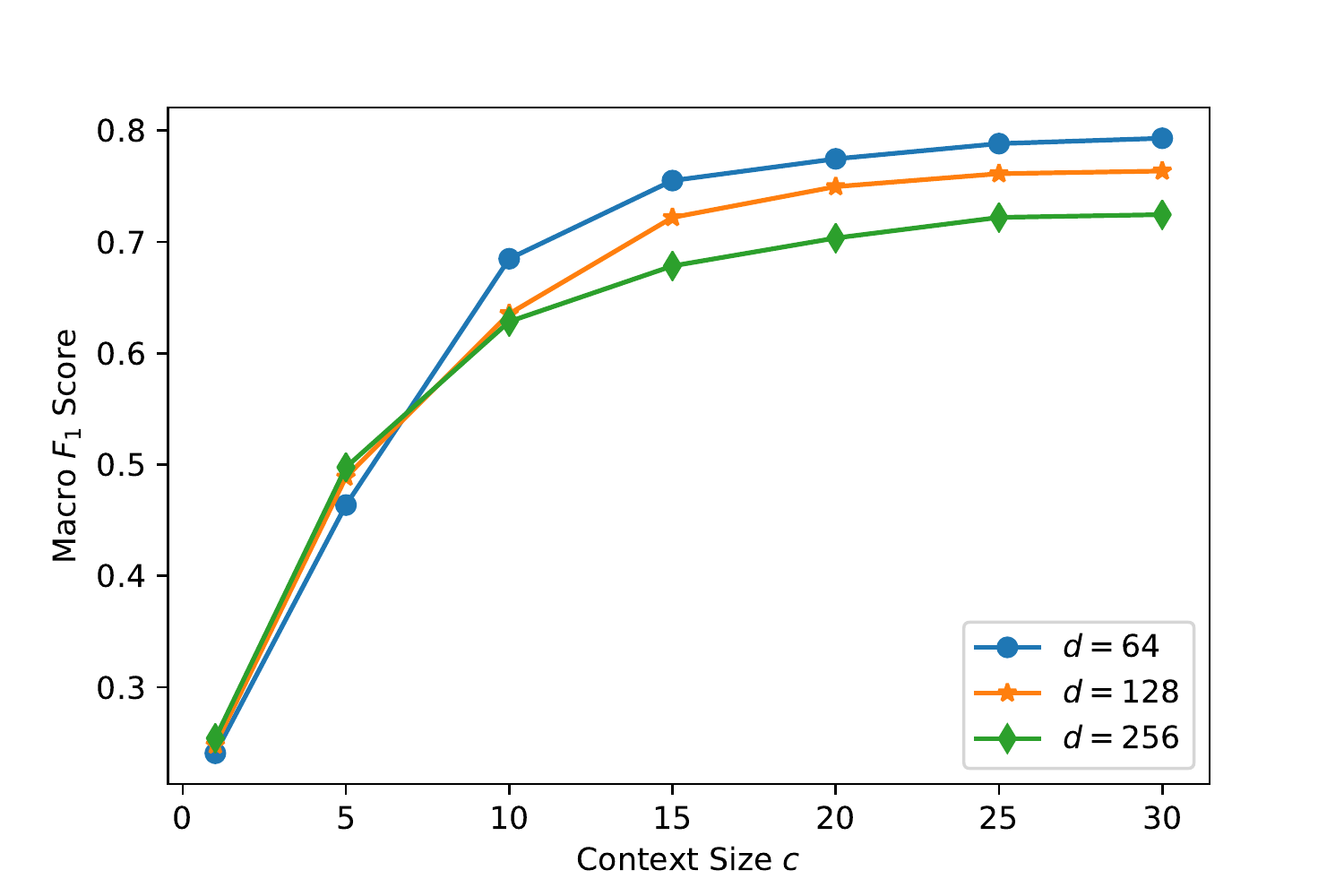}
  \caption{Classification performance at different context sizes on the speed limit classification task using a random forest classifier.\label{fig:context-size-importance}}
\end{figure}

We plot the performance of the random forest classifier on speed limit classification at different context sizes in \cref{fig:context-size-importance}.
As the figure shows, the performance is initially low with a context size of $c=1$, but quickly increases as $c$ increases until performance starts to flatten at $c=15$.
We observe the same trend on road category classification.

Although our results indicate that larger context sizes yield superior or equivalent performance, we expect that very large context sizes can introduce noise.
This could happen if the context size is sufficiently large that otherwise different intersections (and road segments) in different parts of the network have large overlaps between their neighborhoods.
From the perspective of the optimization problem in \cref{eq:network-embedding-objective}, this renders them nearly indistinguishable.

Dimensionality does not influence our conclusions regarding the impact of the other node2vec parameters on classification performance. 
For the random forest classifier, the lowest dimensionality $d=64$ performs up to $20\%$ better than $d=128$ and $d=256$ depending on the choice of $p$ and $q$, as shown in \cref{fig:p-q-performance}. $d=64$ is also the highest performing dimensionality of $d$ at different context sizes, except at a context size of $c=5$, as shown in \cref{fig:context-size-importance}. We suspect this is because the random forest is more prone to overfit the training set for larger values of $d$.

\subsection{Homophily and Classification Performance}
As shown in \cref{tab:data-set-statistics}, there is a skew in the class homophilies for road categories. In particular, the category ''Motorway Approach/Exit'' exhibits substantially lower homophily than the other road categories. We observe that speed limits exhibit both higher homophily on average and higher classification performance in our experiments.
As discussed in \cref{sec:network-embedding}, the network embeddings assume homophily in the network and this assumption is the primary driver for their success.
We therefore expect that the random forest classifier achieves higher performance on road categories such as ''Residential'' that exhibits strong homophily than road categories such as ''Motorway Approach/Exit'' that exhibits weaker homophily.

\begin{table}[]
  \caption{Homophily and $F_1$ scores for each road category using the best random forest classifier.\label{tab:category-homophily-score-relation}}
  \centering
\begin{tabular}{lll}
\toprule
  \emph{Class}   & \emph{Homophily} & \emph{$F_1$ Score} \\
\midrule
  \textbf{Residential}    & $\mathbf{90.4\%}$      & $\mathbf{0.83}$       \\
Trunk          & $81.7\%$      & $0.67$       \\
Motorway       & $78.2\%$      & $0.62$       \\
Unclassified   & $78.0\%$      & $0.62$       \\
  \textbf{Service}        & $\mathbf{72.7\%}$      & $\mathbf{0.56}$       \\
  \textbf{Primary}        & $\mathbf{71.7\%}$      & $\mathbf{0.57}$       \\
Secondary      & $70.6\%$      & $0.54$       \\
Tertiary       & $70.2\%$      & $0.52$       \\
  \textbf{Motorway Approach/Exit} & $\mathbf{36.8\%}$      & $\mathbf{0.25}$       \\
\bottomrule
\end{tabular}
\end{table}

As shown in \cref{tab:category-homophily-score-relation}, there is a near-perfect correspondence between the road categories as ordered by their homophily and macro $F_1$ scores. ''Residential'' and ''Motorway Approach/Exit'' has the highest and lowest $F_1$ scores of $0.83$ and $0.25$, respectively. The only discrepancy is between the ''Service'' and ''Primary'' categories, where ''Primary'' has slightly higher $F_1$ score ($0.57$) than ''Service'' ($0.56$), despite having a slightly lower homophily. This discrepancy may be due to randomness in the training process or disparity between the notion of neighborhoods employed in the homophily measure: we measure homophily based only on the immediate neighbors of a road segment, but the embedding is trained using random walks to represent the node neighborhoods that do not correspond to immediate neighbors. We observe the same pattern for speed limits.

In conclusion, the results in \cref{tab:category-homophily-score-relation} suggests that the classification performance depends strongly on the homophily in the network and that classifier performance is skewed in favor of classes that exhibit strong homophily.

\section{Discussion, Conclusion, and Future Work}\label{sec:conclusion}
We have investigated the suitability of network embedding methods for machine learning on information-sparse road networks by evaluating an existing network embedding method, node2vec, for road category and speed limit classification in the Danish road network.

We have shown that it is possible to achieve macro $F_1$ scores of $0.57$ and $0.79$ on road category classification and speed limit classification, respectively.
Depending on the task, these scores are between $8.3$ and $11.5$ times higher than guessing the most frequent class in the training set.
Furthermore, our results suggests that this performance can be increased further by appropriate parameter tuning of both node2vec and the used classifier.
This suggests that network embedding methods may be useful at extracting structural information from not only social networks, but also road networks.

Some degree of linear separability is implied by the prolific use of linear classifiers in the network embedding literature.
We therefore investigated linear separability in the embedding space for our tasks. 
For both classification tasks, we found that we were unable to fit the training set or generalize to the test using a linear classifier.
By visualizing the embedding space, we found that the members of each class are distributed across several scattered clusters in the embedding space, which reflects the geometric distribution of classes in the network, and suggests that it is difficult to find good linear decision boundaries in the embedding space.

We explored the classification performance for different node2vec parameter configurations.
We found that the impact of the return parameter $p$ and in-out parameter $q$ on classification performance suggests that structural equivalence, as opposed to homophily, is the more appropriate type of similarity.
We also evaluated the other parameters, and found that the dimensionality of the embeddings becomes increasingly more important at high values of $p$ and low values of $q$,
with a preference for low dimensionality for the non-linear classifier and a preference for high dimensionality for the linear classifier.
In addition, it is preferable to include more nodes in the node2vec neighborhood function by using large context sizes, although we expect that large context sizes introduce noise in the embeddings.

Finally, we investigated the relationship between the class homophily and classification performance.
We found that classification performance is better on classes with high homophily than classes with low homophily, and which we propose is a result of node2vec's neighborhood preservation in the embedding space and the corresponding homophily assumption.

We expect that our findings generalize to other network embedding methods.
Specifically, node2vec is a neighborhood preserving network embedding method, and, as demonstrated in our experiments, it tends to achieve higher performance on tasks and classes with higher homophily.
As discussed in \cref{sec:network-embedding}, network embedding methods preserve node neighborhoods in the embedding space as either the main objective or as a sub-objective in combination with attribute information.
We therefore expect that we can achieve similar or better results using different network embedding methods.

The tasks that we have examined exhibit strong homophily. 
In addition, the classes in our data set are scattered in clusters in different areas of the road network.
We therefore expect that our findings generalize to tasks that are similar w.r.t.\ homophily and data distribution in the network.
For instance, driving speeds correlate between adjacent segments~\citep{trajectory-regression}, and driving speeds in, e.g., two distant cities are often similar.

We propose that further attention should given to embedding methods for road networks with sparse information.
To the best of our knowledge, no network embedding method exists that can leverage the spatial information of road networks.
In our data set, road segments may be as short as a few meters for road segments in roundabouts and as a long as several kilometers for motorways.
This means that, e.g., the context size has different meaning for different intersections.
Future work in network embedding should therefore explore the utilization of the spatial information.

We also found that it is hard to predict the road category of motorway approaches and exits due to their low homophily.
This particular road category has low homophily because road segments of this category connect different categories of road segments, e.g., motorways to highways, or vice versa.
Thus, motorway approaches and exits play a particular role in the road network, and structural equivalence as a notion of similarity may be much better suited for this road category. On the other hand, residential road segments are strongly homophilic and appear much better suited for homophily as a notion of similarity.
This suggests that there is a need for road network embedding methods that can capture both structural equivalence and homophily, and balance these notions of similarity for each class.

\section*{Acknowledgments}
This research was supported in part by the DiCyPS project and by grants from the Obel Family Foundation and the Villum Foundation.
We thank the \glsfirst{osm} contributors, without whom this work would not have been possible.
Map data copyrighted \glsdesc{osm} contributors and available from {\url{https://www.openstreetmap.org}}.

  \printbibliography
\end{document}